\pdfoutput=1

\documentclass[11pt]{article}

\usepackage{acl}

\usepackage{times}
\usepackage{latexsym}

\usepackage[T1]{fontenc}

\usepackage[utf8]{inputenc}

\usepackage{microtype}

\usepackage{booktabs} %
\usepackage{subcaption}
\usepackage{marginnote}
\usepackage{bbm}
\usepackage{float}
\usepackage{ifthen}
\usepackage{stmaryrd}
\usepackage{caption}
\usepackage{soul}
\usepackage{pifont}
\usepackage{nicefrac}
\usepackage{cancel}

\newif\ifcomments
\commentstrue
\ifcomments
    \providecommand\abhi[1]{\textcolor{teal}{[AA: #1]}}
    \providecommand\nitish[1]{\textcolor{violet}{[NG: {#1}]}}
    \providecommand\partha[1]{\textcolor{olive}{[PT: #1]}}
    \providecommand\shachi[1]{\textcolor{purple}{[SD: #1]}}
    \providecommand\bidisha[1]{\textcolor{blue}{[BS: #1]}}
    \providecommand\sunita[1]{\textcolor{cyan}{[SS: #1]}}
\else
    \providecommand{\abhi}[1]{}
    \providecommand{\nitish}[1]{}
    \providecommand{\partha}[1]{}
    \providecommand{\shachi}[1]{}
    \providecommand{\bidisha}[1]{}
    \providecommand\sunita[1]{}
\fi

\title{Bootstrapping Multilingual Semantic Parsers \\using Large Language Models}

\newcommand{\aspace}{\hspace{2.0em}}
\newcommand{\google}{$^2$}
\newcommand{\iitb}{$^1$}

\author{Abhijeet Awasthi\iitb\thanks{~~Work done during an internship at Google Research} \aspace 
        Nitish Gupta\google \aspace 
        Bidisha Samanta\google \aspace \\
        \textbf{Shachi Dave\google \aspace 
        Sunita Sarawagi\iitb \aspace  
        Partha Talukdar\google} \\
        \\
        \iitb Indian Institute of Technology Bombay, \google Google Research India  \\
        \texttt{\{awasthi,sunita\}@cse.iitb.ac.in} \\
        \texttt{\{guptanitish,bidishasamanta,shachi,partha\}@google.com} 
        }
  
\usepackage{inconsolata}
\usepackage{amsmath}
\usepackage{graphicx}
\usepackage{adjustbox}
\RequirePackage[textsize=scriptsize,color=yellow!20,
linecolor=orange,bordercolor=orange]{todonotes}
\RequirePackage[inline,shortlabels]{enumitem}
\usepackage{multirow}

\newcommand{\massive}{{\textsc MASSIVE}}
\newcommand{\mtop}{\textsc{MTOP}}
\newcommand{\matisplus}{{\textsc MultiAtis++}}
\newcommand{\mtfive}{mT5}
\newcommand{\mlarge}{Large}

\newcommand{\palm}{\text{PaLM}}

\newcommand{\taf}{\textsc{TaF}}
\newcommand{\gold}{\text{Gold}}

\newcommand{\ours}{{\textsc LLM-T}}
\newcommand{\zeroshot}{{Zero-Shot}}
\newcommand{\fewshot}{{Few-Shot}}
\newcommand{\trainset}{\mathcal{D}}
\newcommand{\seedset}{\mathcal{S}}
\newcommand{\english}{\text{eng}}
\newcommand{\target}{\text{tgt}}

\newcommand{\topk}{\text{top-$k$}}
\newcommand{\topp}{\text{top-$p$}}
\newcommand{\Topk}{\text{Top-$k$}}
\newcommand{\Topp}{\text{Top-$p$}}

\newcommand{\intent}[1]{{\color{magenta} IN:#1}}
\newcommand{\slot}[1]{{\color{cyan} SL:#1}}

\newcommand{\average}{\text{Avg}}

\begin{document}
\maketitle
\begin{abstract}

Despite cross-lingual generalization demonstrated by pre-trained multilingual models, the translate-train paradigm of transferring English datasets across multiple languages remains to be a key mechanism for training task-specific multilingual models. However, for many low-resource languages, the availability of a reliable translation service entails significant amounts of costly human-annotated translation pairs. Further, translation services may continue to be brittle due to domain mismatch between task-specific input text and general-purpose text used for training translation models. For multilingual semantic parsing, we demonstrate the effectiveness and flexibility offered by large language models (LLMs) for translating English datasets into several languages via few-shot prompting. Through extensive comparisons on two public datasets, MTOP and MASSIVE, spanning 50 languages and several domains, we show that our method of translating data using LLMs outperforms a strong translate-train baseline on 41 out of 50 languages. We study the key design choices that enable more effective multilingual data translation via prompted LLMs.

\end{abstract}

\section{Introduction}
Enabling language technologies across several languages is an important goal for serving a diverse range of users in an inclusive manner. Recent advances in large-scale self-supervised multilingual language models 
hold immense promise in bridging the quality gap that currently exists between English and many other low resource languages~\cite{xlmr-conneau-etal-2020-unsupervised, gpt3, mt5-xue-etal-2021-mt5}. Even though multilingual models exhibit cross-lingual generalization, getting meaningful performance across several languages still requires significant amounts of task-specific labeled data.

We consider the problem of automatically synthesizing semantic parsing datasets across several languages.
Semantic parsing~\cite{semp-data-geography-original, zettlemoyerlearning, berant2013semantic} is the task of mapping natural language text into %
an executable {\em logical-form}. 
For example, given a user instruction ($x$) : {\tt ``Wake me up by 5 am''}, mapping it to the logical-form ($y$):~\resizebox{0.43\textwidth}{!}{{\tt [IN:CREATE\_ALARM [SL:DATE\_TIME 5 am ]]}}.
Manual annotation of queries with their logical forms requires human expertise which makes data collection across multiple languages challenging. %

A common approach to automatic multilingual dataset creation is translating existing English datasets into target languages. Prior methods utilize an off-the-shelf machine translation model for translating the English utterance into the target language $x_\english\ \rightarrow x_\target$, followed by projecting language specific components in the English logical-form $y_\english$ to obtain the logical-form $y_\target$ in the target language~\cite{moradshahi-etal-2020-localizing, moradshahi2021contextual, multitop-xia2021multilingual, nicosia2021translate, gritta-etal-2022-crossaligner, wang2022mconala}. The projection step is often learned independent of the translation service, resulting in poor generalization across languages. %

\begin{figure*}[t]
    \centering
    \includegraphics[width=\textwidth]{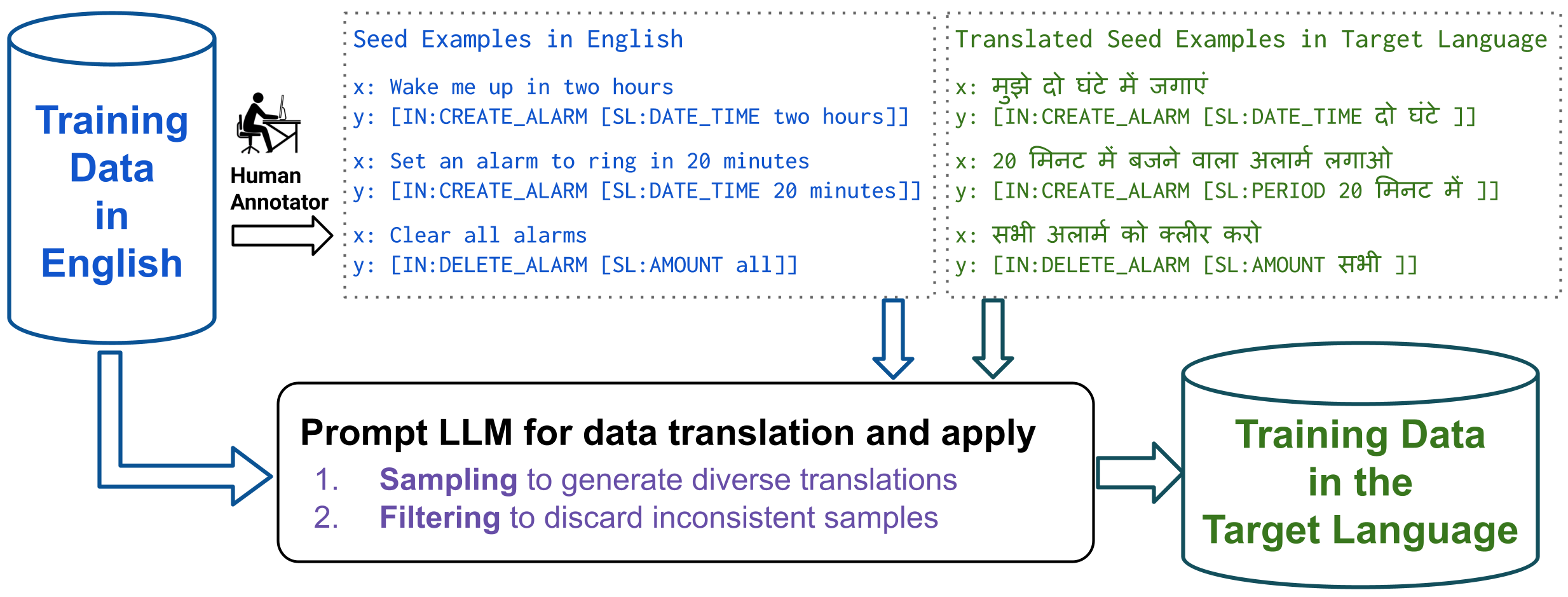}
    \caption{\textbf{Proposed semantic parsing data translation pipeline using LLMs} (\S~\ref{sec:our_method}): With the help of human translators, we first collect translations of a small seed set of English examples in the Target Language (e.g. Hindi; \S~\ref{sec:example_selection}). 
    Given a new English example, a small subset from this initial seed set of examples with their respective translations is chosen to prompt the LLM ~(\S~\ref{sec:prompt_construction}). The prompted LLM translates the given English example in the Target Language. We repeat this process for each example in the English training data to generate a training dataset in the Target Language. To ensure high-quality of the resulting dataset, we generate diverse translations via {\topp} (nucleus) sampling (\S~\ref{sec:sampling}) and apply consistency filtering (\S~\ref{sec:filtering}). %
     }   
    \label{fig:main_figure}
\end{figure*}

In this work we aim to utilize the few-shot generalization abilities exhibited by large language models (LLMs)~\cite{gpt3, chowdhery2022palm, bigscience_workshop_2022}
for bootstrapping semantic parsing datasets across fifty languages. We propose a recipe of using LLMs to translate an English semantic parsing dataset containing {\tt(utterance, logical-form)} pairs: $\trainset_\english = \{(x^i_\english, y^i_\english)\}$ into a corresponding dataset in a target language: $\trainset_\target = \{(x^i_\target, y^i_\target)\}$. 
The generated dataset $\trainset_\target$ is then used to train a semantic parser in the target language. Our method uses a small amount of manually translated semantic parsing examples to \emph{teach} the LLM how to translate English examples in the target language via in-context learning~\cite{min2022rethinking}.

Figure~\ref{fig:main_figure} describes our data-translation pipeline which we refer to as {\ours}~(\S~\ref{sec:our_method}). 
In contrast to prior translation based methods that involved a two-staged process requiring different modules, our method uses the LLM to {\em jointly translate} an English $(x_\english, y_\english)$ pair directly into the target language $(x_\target, y_\target)$.
We identify two important choices that make the LLM translated data more effective for training a downstream parser:
\textbf{(i)~Sampling diverse translations} (\S~\ref{sec:sampling}): %
Decoding translations using \topp{}~\cite{topkfan2018hierarchical} and \topk{}~\cite{toppholtzman2019curious} sampling
leads to improved downstream performance compared to using greedy decoding. %
Sampling multiple diverse translations per example further improves the downstream performance; 
\textbf{(ii)~Filtering inconsistent examples}~(\S~\ref{sec:filtering}): Decoding via sampling can result in noisy joint translations of the (utterance, logical-form) pairs. To filter out the inconsistent pairs, we propose a slot-value match based filtering technique that improves the training data quality. %

We perform experiments on two multilingual semantic parsing datasets: {\mtop}~\cite{mtop-li-etal-2021-mtop} and {\massive}~\cite{fitzgerald2022massive}. On {\em 4 out of 5} languages in {\mtop} and {\em 41 out of 50} languages in {\massive}, our method {\ours} outperforms {\taf}~\cite{nicosia2021translate}, a strong baseline that utilizes a supervised translation service~(\S~\ref{sec:mtop_results}). 
Further, we see that {\ours} achieves 93\% of the performance obtained by ``fully-supervised'' models that use 30$\times$ more manually translated examples (\S~\ref{sec:gold_comparison}).
We justify the importance of generating multiple translations using sampling, filtering out inconsistent examples, and using larger-sized LLMs in improving translated data quality~(\S~\ref{sec:design_choices}). 
Finally, we perform an error analysis of our parser and show the key sources of disagreements between the model predictions and the ground truth~(\S~\ref{sec:error_analysis}).

\section{Background}

\label{sec:background}
In this section, we provide an overview of semantic parsing and prior translation-based methods for creating multilingual semantic parsing datasets.
\subsection{Semantic Parsing} 
Semantic parsing is the task of mapping text queries to their meaning representations or {\em logical forms}~~\cite{semp-data-geography-original, zettlemoyerlearning, berant2013semantic}. 
We focus on task-oriented semantic parsing~\cite{top-gupta2018semantic} 
where the user utterance needs to be parsed into a high-level intent specifying the overall goal, %
and fine-grained slots containing details about the utterance. The intents and slots come from a task-specific vocabulary.
For example, given an utterance $x$: {\tt ``How is the rainfall today?''}, the parser should generate the logical-form \\
$y$:
\resizebox{.95\columnwidth}{!}
{\tt [\intent{GET\_WEATHER} [\slot{ATTRIBUTE} rainfall]}
{\tt[\slot{DATE} today ] ]}

Here, {\tt \intent{GET\_WEATHER}} is the high-level intent, {\tt \slot{ATTRIBUTE}} and {\tt \slot{DATE}} are the slots that specify details about the intent. 
We refer to 
the logical-form with its slot values removed as its "signature". For example, the signature of 
$y$ 
is \\\resizebox{0.48\textwidth}{!}{\tt [\intent{GET\_WEATHER} [\slot{ATTRIBUTE}][\slot{DATE}]]} 

\subsection{Translating Semantic Parsing Datasets}
\label{sec:taf}
Given an English semantic parsing dataset containing (utterance, logical-form) pairs ${\trainset}_{\english}=\{(x^i_{\english},y^i_{\english})\}$, many methods aim to translate ${\trainset}_{\english}$ to a dataset ${\trainset}_{\target}=\{(x^i_{\target},y^i_{\target})\}$ in the target language ({\target}). 
Here $x^i_{\target}$ is the translation of 
$x^i_{\english}$, and $y^i_{\target}$ is the logical form grounded in the translated utterance $x^i_{\target}$.
Target logical form $y^i_{\target}$~has the same signature as $y^i_{\english}$ and only differs in terms of the translated slot values. 
Most translation based approaches~~\cite{moradshahi-etal-2020-localizing, moradshahi2021contextual, multitop-xia2021multilingual, nicosia2021translate} translate an English example $(x^i_{\english}, y^i_{\english})$ to the corresponding target language example $(x^i_{\target}, y^i_{\target})$ via a two step process:
(i)~\textbf{Translate}: Use a supervised translation service to convert the English utterance $x^i_{\english}$ into the target language utterance $x^i_{\target}$; and
(ii)~\textbf{Project}: Replace the English slot values in $y^i_\english$ with spans copied from the translated utterance $x^i_{\target}$ via a learned alignment model. The translated examples are then used to train a downstream multilingual semantic parser. %
For example, \citet{nicosia2021translate} implement the project step by training a filler module on English data to fill slot-values in a logical-form signature by copying spans from the utterance. During inference, the trained filler module is then used in a zero-shot manner to fill logical-form signatures with spans copied from the translated utterances.

\section{Our Method: Prompting LLMs for Dataset Translation}
\label{sec:our_method}
Our goal is to learn a multilingual semantic parser capable of parsing user queries in many languages. Towards this goal, we propose a method for generating multilingual training datasets via few-shot prompting of an LLM to translate existing English datasets into several languages.

In contrast to prior approaches, we jointly perform example translation by prompting an LLM with a few exemplars of translating English $(x_\english, y_\english)$ pairs to target language $(x_\target, y_\target)$ pairs.
Figure~\ref{fig:main_figure} describes our data-translation method which we refer to as {\ours}.
With the help of human translators we first collect a small seed set of exemplar translations 
used for prompting the LLM (\S~\ref{sec:example_selection}). Given an input English example, we dynamically construct the LLM prompt by identifying a relevant subset of seed exemplars (\S~\ref{sec:prompt_construction}). The LLM translates the English example into the target language by in-context learning from the exemplars provided in the prompt. 
Instead of decoding the most likely translation, we generate multiple diverse translations (\S~\ref{sec:sampling}) using {\topp} (nucleus) sampling~\cite{toppholtzman2019curious}.
While sampling improves the text diversity, it can lead to more noisy generations. 
We filter out the noisy generations using a simple string-match based technique before training a parser on the translated data~(\S~\ref{sec:filtering}).

\subsection{Selecting Seed Exemplars for Translation}
\label{sec:example_selection}
Given an English semantic parsing dataset $\trainset_\english=\{(x^i_\english,y^i_\english)\}$, we first want to identify a small seed set $\seedset_\english \subset \trainset_\english$ that will be translated into the target language ($\seedset_\target$) with the help of human translators.
The examples in $\seedset_\english$ and their corresponding translations in $\seedset_\target$ will be used for prompting the LLM.
Therefore, the choice of the seed examples in $\seedset_\english$ that are manually translated into $\seedset_\target$ becomes important---we would like that the multiple domains (e.g. {\tt Alarms, Music, News, Weather, etc.}) and the intents and slot types in each domain are covered. %
This ensures that for a given English example to be translated, we will be able to prompt the LLM in a manner such that at least one of the few-shot exemplars will share the intent and slots with the test English example.
In practice, we select seed examples in a manner to cover all the intents and slots in a domain at least once. If the selected examples are less than 20 for a domain, we select the remaining examples randomly.

\subsection{Constructing the Prompt using Translation Pairs in the Seed Sets}
\label{sec:prompt_construction}
\begin{figure}[t]
    \centering
    \includegraphics[width=\columnwidth]{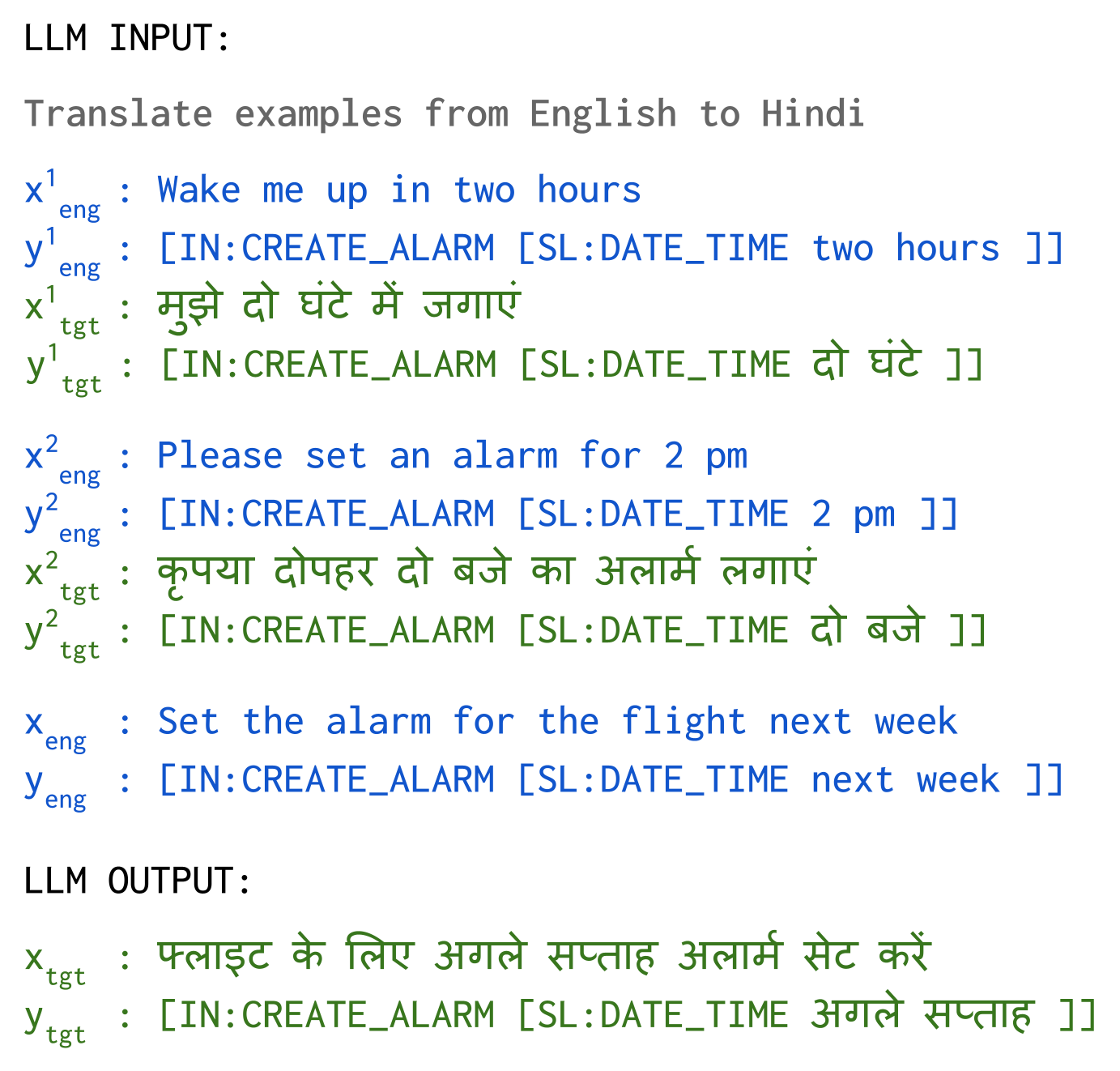}
    \caption{\textbf{Constructing the LLM Prompt} (\S~\ref{sec:prompt_construction}): The input to the LLM contains a brief task description in the beginning followed by a series of English  examples $(x^s_\english, y^s_\english)$ and their translations in the target language $(x^s_\target, y^s_\target)$ chosen from the seed sets $\seedset_\english$ and $\seedset_\target$ respectively. Following the prompt examples, we append the new English example $(x_\english, y_\english)$ to the input prompt which is fed to LLM. In the output, the LLM generates the translation for the new English example $(x_\target, y_\target)$. 
    }
    \label{fig:prompt}
\end{figure}

LLM inference is constrained by the maximum number of tokens in the input. Hence, we can only fit a limited number of examples to construct the LLM prompt. The choice of prompt examples and their ordering is known to significantly impact the quality of the generations~\cite{kumar2021reordering,rubin2021learning,lu2022fantastically}.
To improve the likelihood of correctly translating an English example $(x_\english, y_\english)$, we retrieve seed examples $\{(x^s_\english, y^s_\english, x^s_\target, y^s_\target)\}$ 
that share the same domain with $y_\english$. %
To bias the LLM further, we order the more relevant prompt examples closer to the input English example. Here, relevance between two examples is considered higher if they share the same intent.
The remaining examples are arbitrarily arranged to appear earlier in the prompt. 
Figure~\ref{fig:prompt} shows an example translation---the \emph{LLM input} contains two exemplars and then the English example that needs to be translated. The \emph{LLM output} shows the translated output from the LLM.

\subsection{Decoding Diverse Outputs from LLM}
\label{sec:sampling}
The text decoded from language models using the standard greedy decoding or beam search is often repetitive~\cite{vijayakumar2016diverse, shao2017generating}.  
To mimic how users express the same intentions in diverse ways, we experiment with the {\topk} and {\topp} sampling techniques~\cite{topkfan2018hierarchical, toppholtzman2019curious} to decode multiple diverse translations per example.
We expect sampling multiple translations to yield a better quality training dataset which in turn should result in better downstream semantic parsing performance compared to training on greedily decoded examples.

\subsection{Data Filtering using Slot-Consistency}
\label{sec:filtering}
\begin{figure}[t]
    \centering
    \includegraphics[width=\columnwidth]{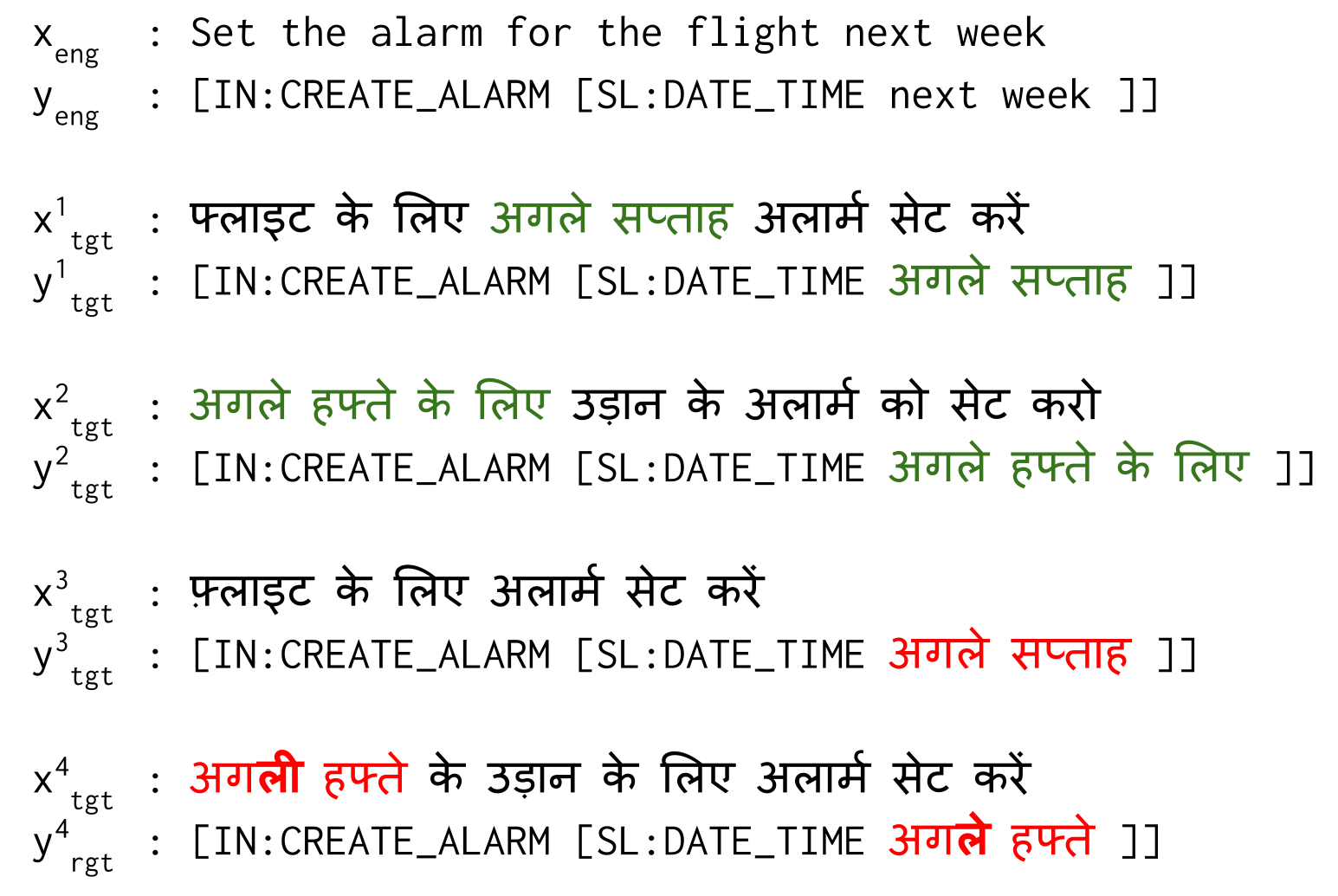}
    
    \caption{\textbf{Slot Consistency Based Filtering} (\S~\ref{sec:filtering}): We present the input English example $(x_\english, y_\english)$ and its four translated samples $\{(x^i_\target, y^i_\target)\})$ the target language. The first two samples are {\em slot-consistent} as the slot-values (in green) in the logical forms appear exactly in the text utterances, while the last two samples are {\em slot-inconsistent} as the slot-values (in red) do not appear as an exact sub-string of the text utterance.
   }
    \label{fig:slot_consistency}
\end{figure}

While the sampling techniques produce more diverse text, the sampled translations can be relatively noisy if they have lower likelihoods as per the model~\cite{zhang2021trading}. Thus, the translated pairs $(x_\target, y_\target)$ in the LLM output can be inconsistent w.r.t. each other. For example, consider the LLM translated pair $(x^3_\target, y^3_\target)$ shown in Figure~\ref{fig:slot_consistency}. Here, $y^3_\target$ contains a slot value (in red) that does not appear in the corresponding utterance $x^3_\target$ making the pair $(x^3_\target, y^3_\target)$ inconsistent.
As per the task definition, for a given example $(x,y)$, the slot-values in the logical form $y$ should come from the spans of the utterance $x$. Thus, we filter out the translated examples $(x_\target, y_\target)$ like these where the slot-values in $y_\target$ do not appear {\em exactly} as an exact sub-span in $x_\target$.
Figure~\ref{fig:slot_consistency} shows examples of slot-consistent and slot-inconsistent generations by an LLM through {\topk} sampling.

\section{Experimental Set-up}
\label{sec:experiments}
We describe our experimental setup in this section.

\paragraph{Datasets} We experiment on two public datasets --- {\mtop}~\cite{mtop-li-etal-2021-mtop} and {\massive}~\cite{fitzgerald2022massive}. %
{\mtop} contains examples from six languages: English, French, German, Hindi, Spanish, and Thai,
spanning 11 domains covering 117 intents and 78 slot types. On average, {\mtop} contains 12.3K examples in the train split, 1.5K in the dev split, and 2.7K in the test split per language. 
{\massive} contains examples from 51 typologically diverse languages including English spanning 18 domains covering 60 intents and 50 slot types. For each language, MASSIVE contains roughly 11.5K examples in the train split, 2K examples in the dev split and 3K examples in the test split.

\paragraph{Evaluation Metric} Prior work~\cite{mtop-li-etal-2021-mtop,nicosia2021translate} uses Exact Match (EM) accuracy as a primary metric which compares predicted and gold logical-forms strings. However, the exact string-match penalizes correct predictions where the order of slots within an intent is different. For example, consider the following logical-forms: \\
{\small {\textbf{LF-1}}:~{\tt[\intent{GET\_WEATHER} [\slot{ATTRIBUTE} rainfall] [\slot{DATE} today ] ]}  \\
\textbf{LF-2}:~{\tt [\intent{GET\_WEATHER} [\slot{DATE} today][\slot{ATTRIBUTE} rainfall ] ]}} \\LF-1 and LF-2 are equivalent but the difference in the ordering of slots results in a negative match.  Thus, we correct the EM metric by making the match function agnostic to the ordering of slots within an intent in the logical-form. We compare different models as per this corrected EM metric.

\paragraph{Semantic Parsing Model}
We use a pre-trained {\mtfive}-{\mlarge} checkpoint (1.2B parameters) to initialize the downstream semantic parsing models that map utterances in the input to logical-forms in the output.
We finetune the \mtfive{} model on the original English dataset mixed with the translated datasets in target languages.
We train using the Adafactor optimizer~\cite{shazeer2018adafactor} with a fixed learning rate of %
$1\mathrm{e}{-3}$ 
and a batch size of 256, for 30K steps using the T5X library~\cite{roberts2022scaling} on 64 TPU-v3 chips. 
Examples from each language are sampled uniformly for batch creation.
For model selection, we choose the best performing checkpoint as per the dev splits and report our results on the test splits. %

\paragraph{\ours{} (Our Method)}
We experiment with 8B, 62B, and 540B sized variants of {\palm}~\cite{chowdhery2022palm} as our LLM, and primarily utilize LLM-540B for translating English examples in different languages. For the seed set $\seedset_\target$ used for prompting the LLM, we borrow roughly 250 examples covering 11 domains from {\mtop}'s train set and 350 examples covering 18 domains from {\massive}'s train set (\S~\ref{sec:example_selection}).
During decoding, we sample $8$ translations per example using {\topp} sampling (\S~\ref{sec:sampling}), with $p=0.95$ and temperature scaling $T=0.7$, followed by filtering out slot-inconsistent examples (\S~\ref{sec:filtering}). We present an analysis of our design choices in \S~\ref{sec:design_choices}.

\paragraph{Baselines}
(i)~\textbf{{\zeroshot}}: Train the model only on the English data and evaluate on other languages in a zero-shot manner. %
(ii)~\textbf{{\fewshot}}: In addition to the English training data,
use the seed set of examples $\seedset_\target$ for each language during training.
For {\mtop}, $|\seedset_\target| \approx 250$ and for {\massive}, $|\seedset_\target| \approx 350$.
(iii)~\textbf{\taf{}}: We implement the method from \citet{nicosia2021translate} that uses an off-the-shelf translation service (\S~\ref{sec:taf}) to construct $\trainset_\target$ in all the target languages. We borrow $\trainset_\target$ from \citet{nicosia2021translate} for {\mtop} and from \citet{nicosia2022evaluating} for {\massive}. %

\begin{figure*}[!ht]
    \centering
    \includegraphics[width=\textwidth]{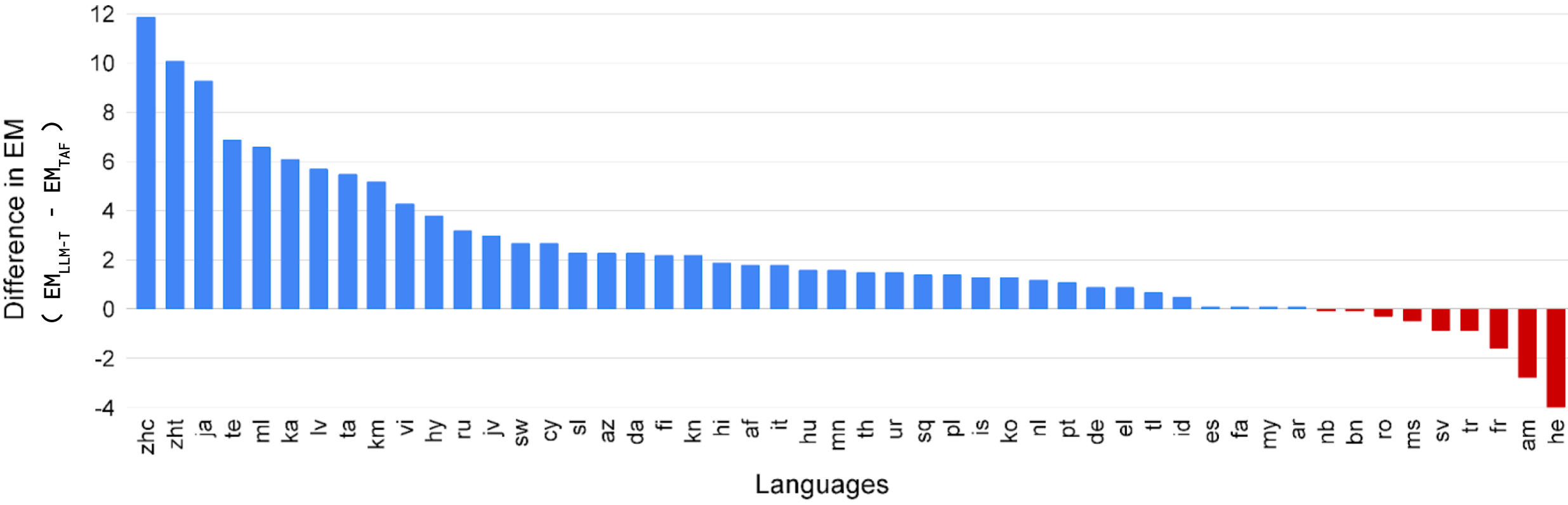}
    \caption{\textbf{EM accuracy difference between {\ours} and {\taf} across the 50 languages in {\massive} dataset} (\S~\ref{sec:mtop_results}). {\ours} outperforms {\taf} on 41 out of 50 languages, with gains of more than $5$ EM for nine of these languages. Only for Hebrew (\texttt{he}), \ours{} performs worse than {\taf} by more than 3 EM. %
    }
    \label{fig:massive_results}
\end{figure*}

\section{Results and Analysis}
\label{sec:main_results}
 We first present downstream performance of semantic parsing models trained on data generated by our method~(\S~\ref{sec:mtop_results}) and compare with zero-shot setting, few-shot setting, and the \taf~method~\cite{nicosia2021translate}. We then compare our method against the ``full-shot'' skyline where we utilize the original training datasets that were manually translated with the help of human annotators in the target languages~(\S~\ref{sec:gold_comparison}). We then present an analysis of different design choices that result in effective data translation using {\ours} (\S~\ref{sec:design_choices}). Finally, we present an error analysis to show the key sources of disagreements between the parser predictions and the ground truth~(\S~\ref{sec:error_analysis}). %
 All the experiments use our corrected EM metric (\S~\ref{sec:experiments}; Evaluation Metric).

\subsection{Evaluation on {\mtop} and {\massive}} 
\label{sec:mtop_results}
In Table~\ref{tab:mtop_results}, we compare performance of different methods for the 5 non-English languages in the {\mtop} dataset. %
The {\zeroshot} baseline trains an {\mtfive} model only on the English part of the train-split. The {\fewshot} baseline additionally includes the human translated seed sets $\seedset_\target$ for each language. %
Both {\taf} and {\ours} train on the original English train set mixed with their respective translated datasets in each language. 
As all the baselines utilize the original English train set, we see comparable performance on English (around $85.0$ EM).  We observe {\ours} outperforms {\taf} in 4 out of 5 languages by $3.6$ EM. Since \ours{} uses $\seedset_\target$ for prompting, we also mix $\seedset_\target$ with \taf{} data and still observe that \ours{} improves over \taf{}+\fewshot{} by $2.9$ EM.
On relatively low-resource languages, Hindi (\texttt{hi}) and Thai (\texttt{th}), \ours{} leads to much larger improvements over \taf{}.

\begin{table}[t]
\centering
\resizebox{\columnwidth}{!}{
\begin{tabular}{l cccccc}
\toprule
\textbf{Method} & \textbf{de} & \textbf{es} & \textbf{fr} & \textbf{hi} & \textbf{th} & \textbf{\average} \\ 
\midrule
    {\zeroshot}   & 54.4    &  57.8    & 62.8   &  42.3    & 42.1 & 51.9 \\
    {\fewshot}  &  62.8    & 69.5   &    65.9 & 55.3    & 53.9   &  61.5 \\
    TAF   & 75.0    &  74.9  &  78.0    &  63.0   & 60.8   &  70.3\\ 
    TAF + {\fewshot} & \textbf{75.1}	&	74.5 &	78.5 &	63.9 &	62.9 & 71.0\\
    \midrule
    {\ours} \small{(ours)}   &  74.0       &  \textbf{75.4}  & \textbf{79.6}   &  \textbf{72.3}    & \textbf{68.0}  & \textbf{73.9} \\
    \bottomrule
\end{tabular}
}
\caption{\textbf{EM accuracy comparison on MTOP} (\S~\ref{sec:mtop_results}): Data generated using {\ours} yields better performance on $4$ out of $5$ languages in MTOP. We observe large improvements for low-resource languages hi and th.
}
\label{tab:mtop_results}
\end{table}

Figure~\ref{fig:massive_results} shows the performance difference between our {\ours} method and {\taf} for the {\massive} dataset~\cite{fitzgerald2022massive}. On 41 out of 50 languages, we find {\ours} to be better than {\taf}. For nine languages {\ours} outperforms {\taf} by more than $5.0$ EM---Simple Mandarin ({\tt zhc}, $+11.9$), Traditional Mandarin ({\tt zht}, $+10.1$), Japanese ({\tt ja}, $+9.3$), Telugu ({\tt te}, $+6.9$), Malayalam ({\tt ml}, $+6.6$), Kannada ({\tt ka}, $+6.1$), Latvian ({\tt lv}, $+5.7$), Tamil ({\tt ta}, $+5.5$), and Khmer ({\tt km}, $+5.2$). Only for Hebrew ({\tt he}, $-4.0$), {\ours} is worse by more than 3.0 EM. 
Averaged across all languages, {\ours} outperforms {\taf} by $2.2$ EM. In Appendix~\ref{sec:app:massive}, we provide detailed baseline comparisons for all the 50 languages.

\subsection{Comparison with gold
translations} 
\label{sec:gold_comparison}
An ideal translate-train method should be competitive w.r.t. training on fully human translated datasets. Table~\ref{tab:gold_comparison} provides a comparison between training on {\taf}, {\ours}, and the datasets fully translated with the help of human annotators in the target languages (\gold{}). %
Between {\taf} and {\gold}, we observe a significant gap of $9.2$~EM in {\mtop} and $6.7$~EM in {\massive}. Our method {\ours}, reduces this gap by $3.6$~EM in {\mtop} and $2.2$~EM in {\massive}. Overall, {\ours} achieves roughly 93\% of the performance obtained by the {\gold} skyline that use more than 30$\times$ human translated examples. Appendix~\ref{sec:app:massive}, provides per-language comparisons with the {\gold} skyline for both the datasets.

\begin{table}[t]
\centering
\small
\begin{tabular}{l cccc}
\toprule
\textbf{Dataset}       & \textbf{{\fewshot}} & \textbf{{\taf}} & \textbf{{\ours}} & \textbf{{\gold}} \\ \midrule
MTOP    &  61.5  &  70.3   & 73.9  & \textbf{79.5}      \\ %
MASSIVE  & 55.9  &  61.0    & 63.2  & \textbf{67.7}    \\ \bottomrule
\end{tabular}
\caption{ 
\textbf{Comparison with {\gold} skyline} (\S~\ref{sec:gold_comparison}): While training on the human translated datasets ({\gold}) yields the best performance, {\ours} results in a smaller performance gap compared to {\taf}. All numbers are averaged over the 5 non-English languages in \mtop{}.
}
\label{tab:gold_comparison}
\end{table}

\begin{table}[ht!]
\centering
\begin{adjustbox}{width=\columnwidth}
\begin{tabular}{c cccccc}
\toprule
\textbf{Decoding} & \multirow{2}{*}{\textbf{de}} & \multirow{2}{*}{\textbf{es}} & \multirow{2}{*}{\textbf{fr}} & \multirow{2}{*}{\textbf{hi}} & \multirow{2}{*}{\textbf{th}} & \multirow{2}{*}{\textbf{\average}} \\ 
\textbf{Strategy} &  &  &  &  & &  \\ \midrule
\addlinespace[1mm]
\multicolumn{1}{l}{\textbf{Greedy}}    &  71.1 &	71.7 &	72.6 &	68.1 &	66.0 &  69.9  \\
\multicolumn{1}{l}{\textbf{\;+ Filtering}}  &  72.2 &	73.5	& 74.8 &	71.5 &	67.4 &	71.9  \\
\midrule
\multicolumn{7}{l}{\textbf{{\Topp} Sampling} ($p=0.95$)} \\  
\small (\#samples)  &  &  &  &  &  &    \\
1 &	70.1  &	71.5 &	74.3 &	66.9 &	67.2 &	70.0 \\
2 &	71.4  &	72.1 &	74.5 &	68.8 &	67.2 &	70.8 \\
4 &	71.1  &	72.8 &	76.4 &	69.0 &	66.0 &	71.1 \\
8 &	71.9  &	72.7 &	74.2 &	70.0 &	\textbf{68.4} &	71.4 \\
\midrule
\multicolumn{7}{l}{\textbf{\Topp{} Sampling} + \textbf{Filtering} ($p=0.95$)} \\
\small (\#samples)  &  &  &  &  &   &  \\
1  &	72.0 &	75.2 &	78.9 &	71.6 &	68.1 &	73.2 \\ 
2  &	73.7 &	75.2 &	79.5 &	72.0 &	67.6 &	73.6 \\ 
4  &	73.4 &	75.3 &	79.0 &	72.1 &	67.7 &	73.5 \\ 
8  &	\textbf{74.0} &	\textbf{75.4} &	\textbf{79.6} &	\textbf{72.3} &	68.0 &	\textbf{73.9} \\ \bottomrule

\end{tabular}
\end{adjustbox}
\caption{\textbf{Impact of decoding strategy and filtering}: Generating multiple translations per English example using {\topp} sampling followed by filtering inconsistent examples offers superior downstream performance compared to using greedy decoding or sampling just one translation per example. In Appendix~\ref{sec:app:decoding} we present results for \topk{} sampling as well.}
\label{tab:decoding_and_filtering}
\end{table}

\subsection{Analysis of Design Choices}
\label{sec:design_choices}
We now present an analysis of the design choices that enabled more effective data translation via {\ours}. All the experiments in this section are carried out on the \mtop{} dataset.

\paragraph{Role of decoding strategy and filtering}
In Table~\ref{tab:decoding_and_filtering}, we present the EM accuracy of parsers trained on datasets translated  using various combinations of decoding (\S~\ref{sec:sampling}) and filtering (\S~\ref{sec:filtering}) methods. For generating the translated outputs we experiment with greedy decoding, {\topk}~\cite{topkfan2018hierarchical} and {\topp}~\cite{toppholtzman2019curious} sampling. Like prior translate-train methods, we begin with only one translation per example and observe %
sampling to be comparable with greedy decoding in downstream EM accuracy. In contrast, decoding two translations per example via %
sampling boosts the EM accuracy across all the languages. However, further increasing the translated samples to $4$ and $8$ results in only marginal performance differences. Manual inspection of the translated data revealed inconsistent utterance and logical-form pairs which motivated our design of slot-consistency based filtering~(\S~\ref{sec:filtering}).
Training the parser %
on filtered data provides further gains over training on unfiltered data. In Appendix~\ref{sec:app:decoding}, we also present the results for \topk{} sampling. Overall, utilizing upto $8$ {\topp} translated samples per English example followed by slot-consistency filtering provides the best performance averaged over all the languages.   

\begin{table}[t]
\centering
\begin{adjustbox}{width=\columnwidth}
\begin{tabular}{c c c c c c c}
\toprule
\textbf{ Max Len}                & \textbf{de} & \textbf{es} & \textbf{fr} & \textbf{hi} & \textbf{th} & \textbf{\average} \\ \midrule

768 &	73.4 &	75.4 &	76.9 &	73.1 &	69.7 &	73.7 \\
1024  &	74.0 &	75.4 &	79.6 &	72.3 &	68.0 &	73.9 \\
1792 &	\textbf{74.3} &	\textbf{75.7} &	\textbf{80.5} &	\textbf{74.0} &	\textbf{71.1} &	\textbf{75.1} \\
\bottomrule
\end{tabular}
\end{adjustbox}
\caption{\textbf{Impact of prompt length}: Longer prompts containing more exemplars result in more effective translated datasets yielding higher EM accuracy.  }
\label{tab:prompt_length}
\end{table}

\paragraph{Impact of Prompt Length}
We expect prompts containing more exemplars to yield higher quality translated examples owing to more information for in-context learning. In Table~\ref{tab:prompt_length}, we compare EM performance when using maximum prompt-lengths of $768, 1024, \text{and} \, 1792$ tokens. Training on datasets translated using prompt-length of 1792 tokens provides the best downstream EM performance across all the languages. However, longer prompts lead to considerably longer inference times. Hence, we conduct our main experiments with %
prompt the length of 1024 tokens.

\paragraph{Role of LLM size} 
In Table~\ref{tab:llm_size}, we compare parser performance when trained on data generated by LLMs of different sizes. Training on larger LLM generated data leads to better performance---{\ours}-540B yields the best performance on all the languages, followed by {\ours}-62B which outperforms {\ours}-8B on all the languages. 
\begin{table}[t]
\centering
\begin{adjustbox}{width=\columnwidth}
\begin{tabular}{l c c c c c c}
\toprule
               & \textbf{de} & \textbf{es} & \textbf{fr} & \textbf{hi} &\textbf{ th} & \textbf{\average} \\ \midrule
{\ours}-8B    & 65.3	&	69.4 &	70.7 &	56.6 &	55.1 & 62.0  \\ 
{\ours}-62B  & 72.0	& 73.3 &	76.7 &	68.2	& 65.6 &	71.2 \\
{\ours}-540B  &	\textbf{74.0} &	\textbf{75.4} &	\textbf{79.6} &	\textbf{72.3} &	\textbf{68.0} &	\textbf{73.9} \\ \bottomrule
\end{tabular}
\end{adjustbox}
\caption{\textbf{Impact of LLM size}: EM performance of semantic parsers trained on translated datasets improve with increasing the size of LLMs used for translation.}
\label{tab:llm_size}
\end{table}

\begin{table*}[!ht]
\centering
\begin{adjustbox}{width=\textwidth}
\begin{tabular}{|l||l|}
\hline
                                     
\multirow{3}{*}{Slot Value Mismatch ($41.1\%$)} & Utterance: Set an alarm for 5 pm tomorrow                                                       \\
                                     & Prediction: {\tt [IN:CREATE\_ALARM [SL:DATE\_TIME {\color{red} for} 5 pm ] [SL:DATE\_TIME tomorrow ]  }                \\
                                     & Target:  {\tt [IN:CREATE\_ALARM [SL:DATE\_TIME 5 pm ] [SL:DATE\_TIME tomorrow ]  }                       \\ \hline\hline
\multirow{3}{*}{Wrong Intent ($19.5\%$)}        & Utterance: What can I do today                                                                             \\
                                     & Prediction:{\tt {[}IN:{\color{red}QUESTION\_NEWS} {[}SL:DATE\_TIME today {]}{]}}                                            \\
                                     & Target: {\tt {[}IN:GET\_EVENT {[}SL:DATE\_TIME today {]} {]}  }                                                  \\ \hline\hline
\multirow{3}{*}{Missing Slot ($15.1\%$)}        & Utterance: Play Justin Timberlake 's newest single                                                         \\
                                     & Prediction:{\tt {[}IN:PLAY\_MUSIC {[}SL:MUSIC\_TYPE single {]} {]}  }                                          \\
                                     & Target: {\tt {[}IN:PLAY\_MUSIC {[}SL:{\color{red}MUSIC\_ARTIST\_NAME Justin Timberlake }{]} {[}SL:MUSIC\_TYPE single {]} {]}} \\ \hline\hline
\multirow{3}{*}{Extra Slot ($14.4\%$)}          & Utterance: play music on the speaker                                                                       \\
                                     & Prediction: {\tt {[}IN:PLAY\_MUSIC {[}SL:MUSIC\_TYPE music {]} {[}SL:{\color{red}MUSIC\_TYPE speaker} {]} {]} }               \\
                                     & Target: {\tt {[}IN:PLAY\_MUSIC {[}SL:MUSIC\_TYPE music {]} {]} }                                                 \\ \hline\hline
\multirow{3}{*}{Slot Confusion ($9.9\%$)}      & Utterance: audio call wedding planner please                                                               \\
                                     & Prediction:{\tt {[}IN:CREATE\_CALL {[}SL:{\color{red} CONTACT} wedding planner {]} {]}      }                                 \\
                                     & Target: {\tt {[}IN:CREATE\_CALL {[}SL:GROUP wedding planner {]} {]} }   
                                     \\ \hline                                      
\end{tabular}
\end{adjustbox}
\caption{\textbf{Examples of Error Categories} (\S~\ref{sec:error_analysis}) The errors in the predicted parse can be broadly classified into five categories:
(i)~Slot Value Mismatch: Predicted parse has the correct signature but the slot-values are incorrect, (ii)~Wrong Intent: High-level intent of the predicted parse is incorrect, (iii)~Missing Slot: One or more slots in the gold parse do not appear in the output, (iv)~Extra Slot: Output contains extra slot(s) compared to the gold,
(v)~Slot Confusion: Prediced parse contains the correct correct intent and number of slots but the wrong slot-types.
}
\label{tab:error_examples}
\end{table*}

\subsection{Error Analysis}
\label{sec:error_analysis}
\begin{figure}[ht!]
    \centering
    \includegraphics[width=0.8\columnwidth]{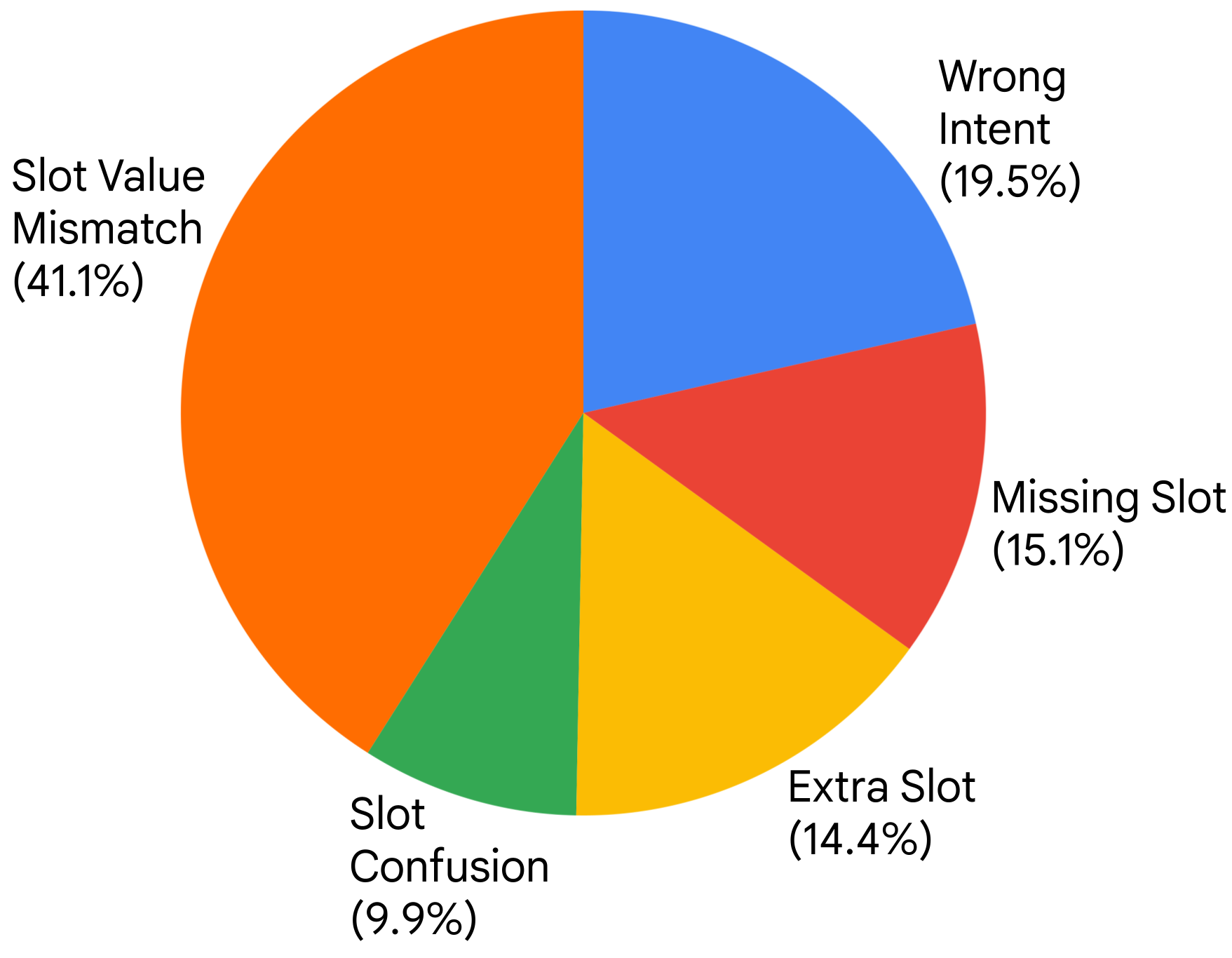}
    \caption{\textbf{Distribution of error categories}:  estimated across all five languages on MTOP's dev set.
    }
    \label{fig:error_dist}
\end{figure}
We analyze the examples where the predictions from our semantic parser do not match with the ground truth.
In Table~\ref{tab:error_examples}, we categorize all the erroneous examples into five broad categories (with English examples): (i)~Slot Value Mismatch (ii)~Wrong Intent (iii)~Missing Slot (iv)~Extra Slot and (v)~Slot Confusion.
Figure~\ref{fig:error_dist} presents the distribution of the error categories aggregated across all the languages on the {\mtop} dev-split.
The "Slot Value Mismatch" is the most frequent error category ($41.1\%$)---here the predicted parse structure is correct but the slot-values do not match perfectly with the gold parse.
After manually inspecting 300 such errors we found that in roughly $50\%$ of the cases the predicted and gold slot-values often have minor mismatches which may not be recognized as error by another human annotator and should not lead to incorrect output upon logical form execution. For example, in the first row of Table~\ref{tab:error_examples}, the predicted value for the {\tt DATE\_TIME} slot is {\tt `for 5 pm'}, while the target value is just {\tt  `5 pm'}.

\section{Related Work}
\label{sec:related_work}
\paragraph{Multilingual Semantic Parsing} Multilingual semantic parsers are typically initialized with a foundation model~\cite{bommasani2021opportunities} pre-trained on vast amounts of multilingual data~\cite{xlmr-conneau-etal-2020-unsupervised, mt5-xue-etal-2021-mt5, mtop-li-etal-2021-mtop, fitzgerald2022massive} followed by supervised training on synthetic or real multilingual datasets. A standard approach for constructing multilingual datasets is to translate and localize English datasets with the help of multilingual speakers or machine translation. For example, {\mtop}~\cite{mtop-li-etal-2021-mtop}, {\massive}~\cite{fitzgerald2022massive}, and {\matisplus}~\cite{multiatisplus} were constructed by translating TOP~\cite{top-gupta2018semantic}, SLURP~\cite{roberts2022scaling}, and ATIS~\cite{atis-price-1990-evaluation} respectively through human translators. 

\paragraph{Machine Translation based methods}
Machine translation based approaches continue to be important for multilingual task-specific models~\cite{hartrumpf2008efficient, transt-liang-etal-2020-xglue, transt-xtreme-pmlr-v119-hu20b, transt-fang2021filter, ladhak-etal-2020-wikilingua} including semantic parsing. Machine translation can either be used during the inference time to translate a user query into English for feeding it to an English-only model. This approach is referred to as {\em translate-test}~\cite{artetxe2020translation, uhrig-etal-2021-translate}. A more common way of using machine translation is in the form of data augmentation, referred as {\em translate-train} where English text in training data is translated into several languages~\cite{sherborne2020bootstrapping, moradshahi-etal-2020-localizing, moradshahi2021contextual, multitop-xia2021multilingual, nicosia2021translate, gritta-etal-2022-crossaligner, wang2022mconala}. In practice, {\em translate-train} methods tend to outperform {\em translate-test} methods while also reducing the latency associated with translating text during the inference time~\cite{yang-etal-2022-learning}.    

\paragraph{LLMs and Few-Shot learning} 
Transformer~\cite{vaswani2017attention} based generative LLMs~\cite{gpt2-radford2019language, gpt3, lamda-thoppilan2022lamda, soltan2022alexatm, megatron-smith2022using, zhang2022opt, chowdhery2022palm}  trained on massive amounts of web-scale text corpora using next token prediction objective exhibit strong few-shot generalization abilities. When prompted with a task description and a handful of task-specific examples, LLMs can often match the performance of finetuned models via in-context learning~\cite{xie2021explanation, min2022rethinking, wei2022chain, zhou2022least}.  We utilize LLMs for translating English datasets in several languages using few-shot prompting.   

\section{Conclusion}
\label{sec:conclusion}
We present a method of utilizing large language models (LLMs) for bootstrapping multilingual semantic parsers across several languages. In comparison to using off-the-shelf translation services that rely on significant amounts of human supervision, we demonstrate that prompting self-supervised LLMs can be a more effective and scalable alternative for dataset translation. We find that generating multiple diverse translations using sampling techniques followed by consistency-based filtering make the translated datasets more effective for training multilingual semantic parsers. %
On 41~out of 50 typologically diverse languages within two large datasets spanning several domains, our method outperforms a strong translate-train method that utilizes a supervised translation service.    %

\section{Limitations}
\label{sec:limitations}
While translating English queries in different languages is a useful form of data augmentation, we think that further performance improvements can be obtained by careful localization of entities in the text queries. This will result in examples where the training dataset contains entities that are often talked about in the target language and might lead to less train-test domain shift. LLMs contain language specific priors which can be harnessed to perform such localization of the translated queries thus enabling more realistic data augmentations. In this work we presented a simple string-match based filtering technique to remove noisy translations. Data filtering can be further improved with the help of learned models. We observed that larger LLMs are important to generate more effective translated data. However running these experiments is constrained by the availability of large amounts of compute resources. We hope future work will address these limitations of our approach.

\section{Ethical Considerations}
We utilize large language models to translate datasets initially available in English into several languages. The real-world deployment of models trained on LLM-translated data should undergo a careful review of any harmful biases. However, the LLM-translated data and the logical-forms generated by a semantic parser are not user-facing, thus a smaller risk of any direct harms. The intended users of any semantic parsing model must be made aware that the answers returned by the model could be incorrect, more so for user-queries in low-resource languages.  We do not immediately foresee any serious negative implications of the specific contributions that we make in this work.

\section{Acknowledgment}
We thank Rahul Goel, Slav Petrov, and Kartikeya Badola for discussions and their feedback on an earlier draft of this paper. We thank Massimo Nicosia for sharing the {\taf} translated datasets and helpful discussions during this project.

\bibliography{main}

\begin{thebibliography}{50}
\expandafter\ifx\csname natexlab\endcsname\relax\def\natexlab#1{#1}\fi

\bibitem[{Artetxe et~al.(2020)Artetxe, Labaka, and
  Agirre}]{artetxe2020translation}
Mikel Artetxe, Gorka Labaka, and Eneko Agirre. 2020.
\newblock Translation artifacts in cross-lingual transfer learning.
\newblock In \emph{Proceedings of the 2020 Conference on Empirical Methods in
  Natural Language Processing (EMNLP)}, pages 7674--7684.

\bibitem[{Berant et~al.(2013)Berant, Chou, Frostig, and
  Liang}]{berant2013semantic}
Jonathan Berant, Andrew Chou, Roy Frostig, and Percy Liang. 2013.
\newblock Semantic parsing on freebase from question-answer pairs.
\newblock In \emph{Proceedings of the 2013 conference on empirical methods in
  natural language processing}, pages 1533--1544.

\bibitem[{Bommasani et~al.(2021)Bommasani, Hudson, Adeli, Altman, Arora, von
  Arx, Bernstein, Bohg, Bosselut, Brunskill
  et~al.}]{bommasani2021opportunities}
Rishi Bommasani, Drew~A Hudson, Ehsan Adeli, Russ Altman, Simran Arora, Sydney
  von Arx, Michael~S Bernstein, Jeannette Bohg, Antoine Bosselut, Emma
  Brunskill, et~al. 2021.
\newblock On the opportunities and risks of foundation models.
\newblock \emph{arXiv preprint arXiv:2108.07258}.

\bibitem[{Brown et~al.(2020)Brown, Mann, Ryder, Subbiah, Kaplan, Dhariwal,
  Neelakantan, Shyam, Sastry, Askell, Agarwal, Herbert-Voss, Krueger, Henighan,
  Child, Ramesh, Ziegler, Wu, Winter, Hesse, Chen, Sigler, Litwin, Gray, Chess,
  Clark, Berner, McCandlish, Radford, Sutskever, and Amodei}]{gpt3}
Tom Brown, Benjamin Mann, Nick Ryder, Melanie Subbiah, Jared~D Kaplan, Prafulla
  Dhariwal, Arvind Neelakantan, Pranav Shyam, Girish Sastry, Amanda Askell,
  Sandhini Agarwal, Ariel Herbert-Voss, Gretchen Krueger, Tom Henighan, Rewon
  Child, Aditya Ramesh, Daniel Ziegler, Jeffrey Wu, Clemens Winter, Chris
  Hesse, Mark Chen, Eric Sigler, Mateusz Litwin, Scott Gray, Benjamin Chess,
  Jack Clark, Christopher Berner, Sam McCandlish, Alec Radford, Ilya Sutskever,
  and Dario Amodei. 2020.
\newblock \href
  {https://proceedings.neurips.cc/paper/2020/file/1457c0d6bfcb4967418bfb8ac142f64a-Paper.pdf}
  {Language models are few-shot learners}.
\newblock In \emph{Advances in Neural Information Processing Systems},
  volume~33, pages 1877--1901. Curran Associates, Inc.

\bibitem[{Chowdhery et~al.(2022)Chowdhery, Narang, Devlin, Bosma, Mishra,
  Roberts, Barham, Chung, Sutton, Gehrmann et~al.}]{chowdhery2022palm}
Aakanksha Chowdhery, Sharan Narang, Jacob Devlin, Maarten Bosma, Gaurav Mishra,
  Adam Roberts, Paul Barham, Hyung~Won Chung, Charles Sutton, Sebastian
  Gehrmann, et~al. 2022.
\newblock Palm: Scaling language modeling with pathways.
\newblock \emph{arXiv preprint arXiv:2204.02311}.

\bibitem[{Conneau et~al.(2020)Conneau, Khandelwal, Goyal, Chaudhary, Wenzek,
  Guzm{\'a}n, Grave, Ott, Zettlemoyer, and
  Stoyanov}]{xlmr-conneau-etal-2020-unsupervised}
Alexis Conneau, Kartikay Khandelwal, Naman Goyal, Vishrav Chaudhary, Guillaume
  Wenzek, Francisco Guzm{\'a}n, Edouard Grave, Myle Ott, Luke Zettlemoyer, and
  Veselin Stoyanov. 2020.
\newblock \href {https://doi.org/10.18653/v1/2020.acl-main.747} {Unsupervised
  cross-lingual representation learning at scale}.
\newblock In \emph{Proceedings of the 58th Annual Meeting of the Association
  for Computational Linguistics}, pages 8440--8451, Online. Association for
  Computational Linguistics.

\bibitem[{Fan et~al.(2018)Fan, Lewis, and Dauphin}]{topkfan2018hierarchical}
Angela Fan, Mike Lewis, and Yann Dauphin. 2018.
\newblock Hierarchical neural story generation.
\newblock In \emph{Proceedings of the 56th Annual Meeting of the Association
  for Computational Linguistics (Volume 1: Long Papers)}, pages 889--898.

\bibitem[{Fang et~al.(2021)Fang, Wang, Gan, Sun, and
  Liu}]{transt-fang2021filter}
Yuwei Fang, Shuohang Wang, Zhe Gan, Siqi Sun, and Jingjing Liu. 2021.
\newblock Filter: An enhanced fusion method for cross-lingual language
  understanding.
\newblock In \emph{Proceedings of the AAAI Conference on Artificial
  Intelligence}, volume~35, pages 12776--12784.

\bibitem[{FitzGerald et~al.(2022)FitzGerald, Hench, Peris, Mackie, Rottmann,
  Sanchez, Nash, Urbach, Kakarala, Singh et~al.}]{fitzgerald2022massive}
Jack FitzGerald, Christopher Hench, Charith Peris, Scott Mackie, Kay Rottmann,
  Ana Sanchez, Aaron Nash, Liam Urbach, Vishesh Kakarala, Richa Singh, et~al.
  2022.
\newblock Massive: A 1m-example multilingual natural language understanding
  dataset with 51 typologically-diverse languages.
\newblock \emph{arXiv preprint arXiv:2204.08582}.

\bibitem[{Gritta et~al.(2022)Gritta, Hu, and
  Iacobacci}]{gritta-etal-2022-crossaligner}
Milan Gritta, Ruoyu Hu, and Ignacio Iacobacci. 2022.
\newblock \href {https://doi.org/10.18653/v1/2022.findings-acl.319}
  {{C}ross{A}ligner {\&} co: Zero-shot transfer methods for task-oriented
  cross-lingual natural language understanding}.
\newblock In \emph{Findings of the Association for Computational Linguistics:
  ACL 2022}, pages 4048--4061, Dublin, Ireland. Association for Computational
  Linguistics.

\bibitem[{Gupta et~al.(2018)Gupta, Shah, Mohit, Kumar, and
  Lewis}]{top-gupta2018semantic}
Sonal Gupta, Rushin Shah, Mrinal Mohit, Anuj Kumar, and Mike Lewis. 2018.
\newblock Semantic parsing for task oriented dialog using hierarchical
  representations.
\newblock In \emph{Proceedings of the 2018 Conference on Empirical Methods in
  Natural Language Processing}, pages 2787--2792.

\bibitem[{Hartrumpf et~al.(2008)Hartrumpf, Gl{\"o}ckner, and
  Leveling}]{hartrumpf2008efficient}
Sven Hartrumpf, Ingo Gl{\"o}ckner, and Johannes Leveling. 2008.
\newblock Efficient question answering with question decomposition and multiple
  answer streams.
\newblock In \emph{Workshop of the Cross-Language Evaluation Forum for European
  Languages}, pages 421--428. Springer.

\bibitem[{Holtzman et~al.(2019)Holtzman, Buys, Du, Forbes, and
  Choi}]{toppholtzman2019curious}
Ari Holtzman, Jan Buys, Li~Du, Maxwell Forbes, and Yejin Choi. 2019.
\newblock The curious case of neural text degeneration.
\newblock In \emph{International Conference on Learning Representations}.

\bibitem[{Hu et~al.(2020)Hu, Ruder, Siddhant, Neubig, Firat, and
  Johnson}]{transt-xtreme-pmlr-v119-hu20b}
Junjie Hu, Sebastian Ruder, Aditya Siddhant, Graham Neubig, Orhan Firat, and
  Melvin Johnson. 2020.
\newblock \href {https://proceedings.mlr.press/v119/hu20b.html} {{XTREME}: A
  massively multilingual multi-task benchmark for evaluating cross-lingual
  generalisation}.
\newblock In \emph{Proceedings of the 37th International Conference on Machine
  Learning}, volume 119 of \emph{Proceedings of Machine Learning Research},
  pages 4411--4421. PMLR.

\bibitem[{Kumar and Talukdar(2021)}]{kumar2021reordering}
Sawan Kumar and Partha Talukdar. 2021.
\newblock Reordering examples helps during priming-based few-shot learning.
\newblock In \emph{Findings of the Association for Computational Linguistics:
  ACL-IJCNLP 2021}, pages 4507--4518.

\bibitem[{Ladhak et~al.(2020)Ladhak, Durmus, Cardie, and
  McKeown}]{ladhak-etal-2020-wikilingua}
Faisal Ladhak, Esin Durmus, Claire Cardie, and Kathleen McKeown. 2020.
\newblock \href {https://doi.org/10.18653/v1/2020.findings-emnlp.360}
  {{W}iki{L}ingua: A new benchmark dataset for cross-lingual abstractive
  summarization}.
\newblock In \emph{Findings of the Association for Computational Linguistics:
  EMNLP 2020}, pages 4034--4048, Online. Association for Computational
  Linguistics.

\bibitem[{Li et~al.(2021)Li, Arora, Chen, Gupta, Gupta, and
  Mehdad}]{mtop-li-etal-2021-mtop}
Haoran Li, Abhinav Arora, Shuohui Chen, Anchit Gupta, Sonal Gupta, and Yashar
  Mehdad. 2021.
\newblock \href {https://doi.org/10.18653/v1/2021.eacl-main.257} {{MTOP}: A
  comprehensive multilingual task-oriented semantic parsing benchmark}.
\newblock In \emph{Proceedings of the 16th Conference of the European Chapter
  of the Association for Computational Linguistics: Main Volume}, pages
  2950--2962, Online. Association for Computational Linguistics.

\bibitem[{Liang et~al.(2020)Liang, Duan, Gong, Wu, Guo, Qi, Gong, Shou, Jiang,
  Cao, Fan, Zhang, Agrawal, Cui, Wei, Bharti, Qiao, Chen, Wu, Liu, Yang,
  Campos, Majumder, and Zhou}]{transt-liang-etal-2020-xglue}
Yaobo Liang, Nan Duan, Yeyun Gong, Ning Wu, Fenfei Guo, Weizhen Qi, Ming Gong,
  Linjun Shou, Daxin Jiang, Guihong Cao, Xiaodong Fan, Ruofei Zhang, Rahul
  Agrawal, Edward Cui, Sining Wei, Taroon Bharti, Ying Qiao, Jiun-Hung Chen,
  Winnie Wu, Shuguang Liu, Fan Yang, Daniel Campos, Rangan Majumder, and Ming
  Zhou. 2020.
\newblock \href {https://doi.org/10.18653/v1/2020.emnlp-main.484} {{XGLUE}: A
  new benchmark datasetfor cross-lingual pre-training, understanding and
  generation}.
\newblock In \emph{Proceedings of the 2020 Conference on Empirical Methods in
  Natural Language Processing (EMNLP)}, pages 6008--6018, Online. Association
  for Computational Linguistics.

\bibitem[{Lu et~al.(2022)Lu, Bartolo, Moore, Riedel, and
  Stenetorp}]{lu2022fantastically}
Yao Lu, Max Bartolo, Alastair Moore, Sebastian Riedel, and Pontus Stenetorp.
  2022.
\newblock Fantastically ordered prompts and where to find them: Overcoming
  few-shot prompt order sensitivity.
\newblock In \emph{Proceedings of the 60th Annual Meeting of the Association
  for Computational Linguistics (Volume 1: Long Papers)}, pages 8086--8098.

\bibitem[{Min et~al.(2022)Min, Lyu, Holtzman, Artetxe, Lewis, Hajishirzi, and
  Zettlemoyer}]{min2022rethinking}
Sewon Min, Xinxi Lyu, Ari Holtzman, Mikel Artetxe, Mike Lewis, Hannaneh
  Hajishirzi, and Luke Zettlemoyer. 2022.
\newblock Rethinking the role of demonstrations: What makes in-context learning
  work?
\newblock \emph{arXiv preprint arXiv:2202.12837}.

\bibitem[{Moradshahi et~al.(2020)Moradshahi, Campagna, Semnani, Xu, and
  Lam}]{moradshahi-etal-2020-localizing}
Mehrad Moradshahi, Giovanni Campagna, Sina Semnani, Silei Xu, and Monica Lam.
  2020.
\newblock \href {https://www.aclweb.org/anthology/2020.emnlp-main.481}
  {Localizing open-ontology {QA} semantic parsers in a day using machine
  translation}.
\newblock In \emph{Proceedings of the 2020 Conference on Empirical Methods in
  Natural Language Processing (EMNLP)}, pages 5970--5983, Online. Association
  for Computational Linguistics.

\bibitem[{Moradshahi et~al.(2021)Moradshahi, Tsai, Campagna, and
  Lam}]{moradshahi2021contextual}
Mehrad Moradshahi, Victoria Tsai, Giovanni Campagna, and Monica~S Lam. 2021.
\newblock Contextual semantic parsing for multilingual task-oriented dialogues.
\newblock \emph{arXiv preprint arXiv:2111.02574}.

\bibitem[{Nicosia and Piccinno(2022)}]{nicosia2022evaluating}
Massimo Nicosia and Francesco Piccinno. 2022.
\newblock Evaluating byte and wordpiece level models for massively multilingual
  semantic parsing.
\newblock \emph{arXiv preprint arXiv:2212.07223}.

\bibitem[{Nicosia et~al.(2021)Nicosia, Qu, and Altun}]{nicosia2021translate}
Massimo Nicosia, Zhongdi Qu, and Yasemin Altun. 2021.
\newblock Translate \& fill: Improving zero-shot multilingual semantic parsing
  with synthetic data.
\newblock In \emph{Findings of the Association for Computational Linguistics:
  EMNLP 2021}, pages 3272--3284.

\bibitem[{Price(1990)}]{atis-price-1990-evaluation}
P.~J. Price. 1990.
\newblock \href {https://aclanthology.org/H90-1020} {Evaluation of spoken
  language systems: the {ATIS} domain}.
\newblock In \emph{Speech and Natural Language: Proceedings of a Workshop Held
  at Hidden Valley, {P}ennsylvania, June 24-27,1990}.

\bibitem[{Radford et~al.(2019)Radford, Wu, Child, Luan, Amodei, and
  Sutskever}]{gpt2-radford2019language}
Alec Radford, Jeff Wu, Rewon Child, David Luan, Dario Amodei, and Ilya
  Sutskever. 2019.
\newblock Language models are unsupervised multitask learners.

\bibitem[{Roberts et~al.(2022)Roberts, Chung, Levskaya, Mishra, Bradbury,
  Andor, Narang, Lester, Gaffney, Mohiuddin et~al.}]{roberts2022scaling}
Adam Roberts, Hyung~Won Chung, Anselm Levskaya, Gaurav Mishra, James Bradbury,
  Daniel Andor, Sharan Narang, Brian Lester, Colin Gaffney, Afroz Mohiuddin,
  et~al. 2022.
\newblock Scaling up models and data with {\tt t5x and seqio}.
\newblock \emph{arXiv preprint arXiv:2203.17189}.

\bibitem[{Rubin et~al.(2021)Rubin, Herzig, and Berant}]{rubin2021learning}
Ohad Rubin, Jonathan Herzig, and Jonathan Berant. 2021.
\newblock Learning to retrieve prompts for in-context learning.
\newblock \emph{arXiv preprint arXiv:2112.08633}.

\bibitem[{Scao et~al.(2022)Scao, Fan, Akiki, Pavlick, Ili{\'c}, Hesslow,
  Castagn{\'e}, Luccioni, Yvon, Gall{\'e} et~al.}]{bigscience_workshop_2022}
Teven~Le Scao, Angela Fan, Christopher Akiki, Ellie Pavlick, Suzana Ili{\'c},
  Daniel Hesslow, Roman Castagn{\'e}, Alexandra~Sasha Luccioni, Fran{\c{c}}ois
  Yvon, Matthias Gall{\'e}, et~al. 2022.
\newblock Bloom: A 176b-parameter open-access multilingual language model.
\newblock \emph{arXiv preprint arXiv:2211.05100}.

\bibitem[{Shao et~al.(2017)Shao, Gouws, Britz, Goldie, Strope, and
  Kurzweil}]{shao2017generating}
Yuanlong Shao, Stephan Gouws, Denny Britz, Anna Goldie, Brian Strope, and Ray
  Kurzweil. 2017.
\newblock Generating high-quality and informative conversation responses with
  sequence-to-sequence models.
\newblock In \emph{Proceedings of the 2017 Conference on Empirical Methods in
  Natural Language Processing}, pages 2210--2219.

\bibitem[{Shazeer and Stern(2018)}]{shazeer2018adafactor}
Noam Shazeer and Mitchell Stern. 2018.
\newblock Adafactor: Adaptive learning rates with sublinear memory cost.
\newblock In \emph{International Conference on Machine Learning}, pages
  4596--4604. PMLR.

\bibitem[{Sherborne et~al.(2020)Sherborne, Xu, and
  Lapata}]{sherborne2020bootstrapping}
Tom Sherborne, Yumo Xu, and Mirella Lapata. 2020.
\newblock Bootstrapping a crosslingual semantic parser.
\newblock In \emph{Findings of the Association for Computational Linguistics:
  EMNLP 2020}, pages 499--517.

\bibitem[{Smith et~al.(2022)Smith, Patwary, Norick, LeGresley, Rajbhandari,
  Casper, Liu, Prabhumoye, Zerveas, Korthikanti
  et~al.}]{megatron-smith2022using}
Shaden Smith, Mostofa Patwary, Brandon Norick, Patrick LeGresley, Samyam
  Rajbhandari, Jared Casper, Zhun Liu, Shrimai Prabhumoye, George Zerveas,
  Vijay Korthikanti, et~al. 2022.
\newblock Using deepspeed and megatron to train megatron-turing nlg 530b, a
  large-scale generative language model.
\newblock \emph{arXiv preprint arXiv:2201.11990}.

\bibitem[{Soltan et~al.(2022)Soltan, Ananthakrishnan, FitzGerald, Gupta, Hamza,
  Khan, Peris, Rawls, Rosenbaum, Rumshisky et~al.}]{soltan2022alexatm}
Saleh Soltan, Shankar Ananthakrishnan, Jack FitzGerald, Rahul Gupta, Wael
  Hamza, Haidar Khan, Charith Peris, Stephen Rawls, Andy Rosenbaum, Anna
  Rumshisky, et~al. 2022.
\newblock Alexatm 20b: Few-shot learning using a large-scale multilingual
  seq2seq model.
\newblock \emph{arXiv preprint arXiv:2208.01448}.

\bibitem[{Thoppilan et~al.(2022)Thoppilan, De~Freitas, Hall, Shazeer,
  Kulshreshtha, Cheng, Jin, Bos, Baker, Du et~al.}]{lamda-thoppilan2022lamda}
Romal Thoppilan, Daniel De~Freitas, Jamie Hall, Noam Shazeer, Apoorv
  Kulshreshtha, Heng-Tze Cheng, Alicia Jin, Taylor Bos, Leslie Baker, Yu~Du,
  et~al. 2022.
\newblock Lamda: Language models for dialog applications.
\newblock \emph{arXiv preprint arXiv:2201.08239}.

\bibitem[{Uhrig et~al.(2021)Uhrig, Garcia, Opitz, and
  Frank}]{uhrig-etal-2021-translate}
Sarah Uhrig, Yoalli Garcia, Juri Opitz, and Anette Frank. 2021.
\newblock \href {https://doi.org/10.18653/v1/2021.iwpt-1.6} {Translate, then
  parse! a strong baseline for cross-lingual {AMR} parsing}.
\newblock In \emph{Proceedings of the 17th International Conference on Parsing
  Technologies and the IWPT 2021 Shared Task on Parsing into Enhanced Universal
  Dependencies (IWPT 2021)}, pages 58--64, Online. Association for
  Computational Linguistics.

\bibitem[{Vaswani et~al.(2017)Vaswani, Shazeer, Parmar, Uszkoreit, Jones,
  Gomez, Kaiser, and Polosukhin}]{vaswani2017attention}
Ashish Vaswani, Noam Shazeer, Niki Parmar, Jakob Uszkoreit, Llion Jones,
  Aidan~N Gomez, {\L}ukasz Kaiser, and Illia Polosukhin. 2017.
\newblock Attention is all you need.
\newblock \emph{Advances in neural information processing systems}, 30.

\bibitem[{Vijayakumar et~al.(2016)Vijayakumar, Cogswell, Selvaraju, Sun, Lee,
  Crandall, and Batra}]{vijayakumar2016diverse}
Ashwin~K Vijayakumar, Michael Cogswell, Ramprasath~R Selvaraju, Qing Sun,
  Stefan Lee, David Crandall, and Dhruv Batra. 2016.
\newblock Diverse beam search: Decoding diverse solutions from neural sequence
  models.
\newblock \emph{arXiv preprint arXiv:1610.02424}.

\bibitem[{Wang et~al.(2022)Wang, Cuenca, Zhou, Xu, and
  Neubig}]{wang2022mconala}
Zhiruo Wang, Grace Cuenca, Shuyan Zhou, Frank~F Xu, and Graham Neubig. 2022.
\newblock Mconala: A benchmark for code generation from multiple natural
  languages.
\newblock \emph{arXiv preprint arXiv:2203.08388}.

\bibitem[{Wei et~al.(2022)Wei, Wang, Schuurmans, Bosma, Chi, Le, and
  Zhou}]{wei2022chain}
Jason Wei, Xuezhi Wang, Dale Schuurmans, Maarten Bosma, Ed~Chi, Quoc Le, and
  Denny Zhou. 2022.
\newblock Chain of thought prompting elicits reasoning in large language
  models.
\newblock \emph{arXiv preprint arXiv:2201.11903}.

\bibitem[{Xia and Monti(2021)}]{multitop-xia2021multilingual}
Menglin Xia and Emilio Monti. 2021.
\newblock Multilingual neural semantic parsing for low-resourced languages.
\newblock In \emph{The Tenth Joint Conference on Lexical and Computational
  Semantics}.

\bibitem[{Xie et~al.(2021)Xie, Raghunathan, Liang, and Ma}]{xie2021explanation}
Sang~Michael Xie, Aditi Raghunathan, Percy Liang, and Tengyu Ma. 2021.
\newblock An explanation of in-context learning as implicit bayesian inference.
\newblock In \emph{International Conference on Learning Representations}.

\bibitem[{Xu et~al.(2020)Xu, Haider, and Mansour}]{multiatisplus}
Weijia Xu, Batool Haider, and Saab Mansour. 2020.
\newblock End-to-end slot alignment and recognition for cross-lingual nlu.
\newblock In \emph{Proceedings of the 2020 Conference on Empirical Methods in
  Natural Language Processing (EMNLP)}, pages 5052--5063.

\bibitem[{Xue et~al.(2021)Xue, Constant, Roberts, Kale, Al-Rfou, Siddhant,
  Barua, and Raffel}]{mt5-xue-etal-2021-mt5}
Linting Xue, Noah Constant, Adam Roberts, Mihir Kale, Rami Al-Rfou, Aditya
  Siddhant, Aditya Barua, and Colin Raffel. 2021.
\newblock \href {https://doi.org/10.18653/v1/2021.naacl-main.41} {m{T}5: A
  massively multilingual pre-trained text-to-text transformer}.
\newblock In \emph{Proceedings of the 2021 Conference of the North American
  Chapter of the Association for Computational Linguistics: Human Language
  Technologies}, pages 483--498, Online. Association for Computational
  Linguistics.

\bibitem[{Yang et~al.(2022)Yang, Parikh, and Raffel}]{yang-etal-2022-learning}
Diyi Yang, Ankur Parikh, and Colin Raffel. 2022.
\newblock \href {https://doi.org/10.18653/v1/2022.acl-tutorials.5} {Learning
  with limited text data}.
\newblock In \emph{Proceedings of the 60th Annual Meeting of the Association
  for Computational Linguistics: Tutorial Abstracts}, pages 28--31, Dublin,
  Ireland. Association for Computational Linguistics.

\bibitem[{Zelle and Mooney(1996)}]{semp-data-geography-original}
John~M. Zelle and Raymond~J. Mooney. 1996.
\newblock \href {http://www.cs.utexas.edu/users/ai-lab?zelle:aaai96} {Learning
  to parse database queries using inductive logic programming}.
\newblock In \emph{AAAI/IAAI}, pages 1050--1055, Portland, OR. AAAI Press/MIT
  Press.

\bibitem[{Zettlemoyer and Collins(2005)}]{zettlemoyerlearning}
Luke~S. Zettlemoyer and Michael Collins. 2005.
\newblock Learning to map sentences to logical form: Structured classification
  with probabilistic categorial grammars.
\newblock In \emph{Proceedings of the Twenty-First Conference on Uncertainty in
  Artificial Intelligence}, UAI'05, page 658–666, Arlington, Virginia, USA.
  AUAI Press.

\bibitem[{Zhang et~al.(2021)Zhang, Duckworth, Ippolito, and
  Neelakantan}]{zhang2021trading}
Hugh Zhang, Daniel Duckworth, Daphne Ippolito, and Arvind Neelakantan. 2021.
\newblock Trading off diversity and quality in natural language generation.
\newblock In \emph{Proceedings of the Workshop on Human Evaluation of NLP
  Systems (HumEval)}, pages 25--33.

\bibitem[{Zhang et~al.(2022)Zhang, Roller, Goyal, Artetxe, Chen, Chen, Dewan,
  Diab, Li, Lin et~al.}]{zhang2022opt}
Susan Zhang, Stephen Roller, Naman Goyal, Mikel Artetxe, Moya Chen, Shuohui
  Chen, Christopher Dewan, Mona Diab, Xian Li, Xi~Victoria Lin, et~al. 2022.
\newblock Opt: Open pre-trained transformer language models.
\newblock \emph{arXiv preprint arXiv:2205.01068}.

\bibitem[{Zhou et~al.(2022)Zhou, Sch{\"a}rli, Hou, Wei, Scales, Wang,
  Schuurmans, Bousquet, Le, and Chi}]{zhou2022least}
Denny Zhou, Nathanael Sch{\"a}rli, Le~Hou, Jason Wei, Nathan Scales, Xuezhi
  Wang, Dale Schuurmans, Olivier Bousquet, Quoc Le, and Ed~Chi. 2022.
\newblock Least-to-most prompting enables complex reasoning in large language
  models.
\newblock \emph{arXiv preprint arXiv:2205.10625}.

\end{thebibliography}

\clearpage
\appendix
\setcounter{table}{0}
\renewcommand{\thetable}{A\arabic{table}}
\section{Appendix}
\label{sec:appendix}
\subsection{Additional results}
\label{sec:app:massive}
\begin{table}[ht!]
\centering
\small
\begin{adjustbox}{width=\columnwidth}
\begin{tabular}{@{}lcccccc|c@{}}
\toprule
\multirow{2}{*}{\textbf{Lang}} & \multirow{2}{*}{\textbf{\zeroshot}} & \multirow{2}{*}{\textbf{\fewshot}} & \multirow{2}{*}{\textbf{\taf}} & \multirow{2}{*}{\begin{tabular}[c]{@{}l@{}}\textbf{\taf} + \\ \textbf{\fewshot}\end{tabular}} & \multirow{2}{*}{\begin{tabular}[c]{@{}l@{}}\textbf{\ours}\\ (\topk{})\end{tabular}} & \multirow{2}{*}{\begin{tabular}[c]{@{}l@{}}\textbf{\ours{}}\\ (\topp{})\end{tabular}} & \multirow{2}{*}{\begin{tabular}[c]{@{}l@{}}\textbf{\gold} \\ (skyline) \end{tabular}} \\
                      &                            &                           &                      &                                                                        &                                                                          &                                                                          &                       \\ \midrule
af   & 48.5      & 59.0       & 64.5 & 64.5        & 66.7        & 66.3        & 68.5 \\
am   & 31.0      & 47.6     & 58.3 & 57.4        & 56.1        & 55.5        & 64.6 \\
ar   & 35.9      & 50.5     & 57.2 & 58.0          & 56.6        & 57.3        & 65.5 \\
az   & 39.3      & 57.1     & 60.5 & 60.5        & 62.6        & 62.8        & 68.8 \\
bn   & 40.8      & 55.4     & 62.1 & 61.7        & 61.1        & 62.0          & 68.3 \\
cy   & 26.7      & 44.8     & 59.1 & 58.3        & 61.1        & 61.8        & 65.1 \\
da   & 57.5      & 62.4     & 66.2 & 66.3        & 69.1        & 68.5        & 71.0   \\
de   & 54.3      & 62.8     & 67.5 & 67.7        & 68.3        & 68.4        & 70.4 \\
el   & 47.3      & 57.8     & 64.2 & 65.5        & 65.2        & 65.1        & 68.7 \\
en	 & 72.7	     & 71.4	    & 73.5 & 72.9	     & 73.3	       & 73.4	     & 73.0  \\
es   & 53.4      & 58.1     & 64.6 & 64.6        & 64.7        & 64.7        & 66.6 \\
fa   & 48.8      & 58.0       & 63.1 & 62.9        & 62.8        & 63.2        & 68.1 \\
fi   & 47.5      & 58.4     & 65.0   & 65.3        & 66.7        & 67.2        & 70.9 \\
fr   & 54.6      & 58.0       & 65.3 & 64.9        & 63.9        & 63.7        & 67.1 \\
he   & 35.3      & 56.1     & 60.6 & 61.2        & 55.3        & 56.6        & 68.3 \\
hi   & 40.1      & 54.4     & 61.6 & 62.5        & 63.1        & 63.5        & 66.2 \\
hu   & 44.1      & 57.1     & 63.8 & 63.6        & 64.5        & 65.4        & 69.7 \\
hy   & 39.3      & 53.8     & 58.7 & 59.2        & 62.3        & 62.5        & 67.1 \\
id   & 55.3      & 60.2     & 65.5 & 65.9        & 66.6        & 66.0          & 69.1 \\
is   & 41.3      & 54.4     & 62.2 & 61.5        & 63.6        & 63.5        & 69.5 \\
it   & 52.3      & 58.6     & 64.0   & 63.6        & 65.2        & 65.8        & 67.2 \\
ja   & 45.6      & 55.1     & 56.3 & 56.5        & 65.6        & 65.6        & 67.3 \\
jv   & 34.3      & 51.7     & 58.6 & 60.2        & 62.0          & 61.6        & 66.7 \\
ka   & 36.5      & 53.4     & 53.5 & 54.6        & 59.2        & 59.6        & 65.7 \\
km   & 37.8      & 51.1     & 49.1 & 53.7        & 55.3        & 54.3        & 62.8 \\
kn   & 37.1      & 49.3     & 55.0   & 55.9        & 57.7        & 57.2        & 62.1 \\
ko   & 42.1      & 56.3     & 62.2 & 63.6        & 62.4        & 63.5        & 69.3 \\
lv   & 45.4      & 56.0       & 60.4 & 61.3        & 66.0          & 66.1        & 68.8 \\
ml   & 38.6      & 53.9     & 55.5 & 56.9        & 62.5        & 62.1        & 67.5 \\
mn   & 30.9      & 51.4     & 57.6 & 59.4        & 59.5        & 59.2        & 68.0   \\
ms   & 48.6      & 58.9     & 66.2 & 66.2        & 65.8        & 65.7        & 69.2 \\
my   & 38.1      & 54.9     & 60.5 & 62.3        & 61.5        & 60.6        & 69.6 \\
nb   & 55.2      & 63.0       & 67.5 & 67.7        & 67.7        & 67.4        & 71.0   \\
nl   & 53.1      & 61.2     & 67.3 & 68.5        & 68.7        & 68.5        & 70.5 \\
pl   & 50.5      & 57.4     & 61.1 & 61.4        & 62.9        & 62.5        & 65.6 \\
pt   & 54.9      & 60.3     & 65.8 & 65.7        & 66.4        & 66.9        & 68.5 \\
ro   & 51.2      & 58.8     & 65.4 & 65.0          & 64.8        & 65.1        & 68.8 \\
ru   & 42.3      & 59.4     & 63.0   & 63.1        & 66.6        & 66.2        & 69.4 \\
sl   & 46.0        & 57.8     & 63.1 & 64.0          & 65.3        & 65.4        & 68.8 \\
sq   & 41.0       & 55.4     & 60.3 & 60.4        & 62.1        & 61.7        & 67.3 \\
sv   & 57.2      & 63.1     & 69.8 & 69.6        & 69.3        & 68.9        & 72.4 \\
sw   & 35.7      & 52.3     & 57.9 & 57.5        & 60.9        & 60.6        & 65.3 \\
ta   & 37.2      & 53.0       & 55.4 & 55.7        & 60.7        & 60.9        & 65.8 \\
te   & 38.7      & 49.0       & 51.6 & 53.6        & 56.8        & 58.5        & 61.6 \\
th   & 49.4      & 60.0       & 63.5 & 66.5        & 65.2        & 65          & 71.5 \\
tl   & 48.4      & 55.7     & 64.1 & 64.2        & 65.2        & 64.8        & 67.5 \\
tr   & 46.7      & 58.5     & 63.7 & 63.4        & 62.7        & 62.8        & 69.4 \\
ur   & 38.9      & 51.2     & 60.4 & 60.6        & 62.2        & 61.9        & 64.6 \\
vi   & 46.9      & 55.1     & 59.0   & 59.2        & 63.0          & 63.3        & 67.6 \\
zhc  & 34.7      & 56.1     & 52.0   & 53.9        & 64.2        & 63.9        & 66.3 \\
zht  & 35.2      & 51.8     & 50.5 & 52.3        & 60.7        & 60.6        & 63.6 \\ \midrule
{\average}  & 43.8      & 55.9     & 61.0 & 61.6        & 63.2        & 63.2        & 67.7 \\ \bottomrule
\end{tabular}
\end{adjustbox}
\caption{EM accuracy comparison on {\massive} dataset. {\average} reports the EM accuracy averaged across the 50 non-English languages}
\label{tab:app:massive_results}
\end{table}

\begin{table}[ht!]
\centering
\small
\begin{adjustbox}{width=\columnwidth}
\begin{tabular}{@{}lcccccc|c@{}}
\toprule
\multirow{2}{*}{\textbf{Lang}} & \multirow{2}{*}{\textbf{\zeroshot}} & \multirow{2}{*}{\textbf{\fewshot}} & \multirow{2}{*}{\textbf{\taf}} & \multirow{2}{*}{\begin{tabular}[c]{@{}l@{}}\textbf{\taf} + \\ \textbf{\fewshot}\end{tabular}} & \multirow{2}{*}{\begin{tabular}[c]{@{}l@{}}\textbf{\ours}\\ (\topk{})\end{tabular}} & \multirow{2}{*}{\begin{tabular}[c]{@{}l@{}}\textbf{\ours{}}\\ (\topp{})\end{tabular}} & \multirow{2}{*}{\begin{tabular}[c]{@{}l@{}}\textbf{\gold} \\ (skyline) \end{tabular}} \\
                      &                            &                           &                      &                                                                        &                                                                          &                                                                          &                       \\ \midrule
de	& 54.4	& 62.8	& 75.0 &	75.1 &	73.7 &	74.0 &	78.5 \\
es	& 57.8	& 69.5	& 74.9 &	74.5 &	75.2 &	75.4 &	82.9 \\
fr	& 62.8	& 65.9	& 78.0 &	78.5 &	79.7 &	79.6 &	80.8 \\
hi 	& 42.3	& 55.3	& 63.0 &	63.9 &	72.5 &	72.3 &	78.5 \\
th	& 42.1	& 53.9	& 60.8 &	62.9 &	66.8 &	68.0 &	77.0 \\
en	& 84.1	& 84.0	& 85.2 &	85.0 &	85.2 &	85.1 &	85.4 \\
\midrule
{\average}	& 51.9	& 61.5	& 70.3 &	71.0 &	73.6 &	73.9 &	79.5 \\ \bottomrule
\end{tabular}
\end{adjustbox}
\caption{EM accuracy comparison on {\mtop} dataset. {\average} reports the EM accuracy averaged across the 5 non-English languages}
\label{tab:app:mtop_results}
\end{table}

In Table~\ref{tab:app:massive_results}, we present detailed baseline comparisons for all the 51 languages in the \massive{} dataset. {\zeroshot}, {\fewshot}, {\taf}, and {\taf}+{\fewshot} are the baselines described in Section~\ref{sec:experiments}. {\ours} represents our method with {\topk} or {\topp} sampling used while decoding the translated examples. {\gold} is the "full-shot" skyline which utilizes the original human-translated datasets (\S~\ref{sec:gold_comparison}). Table~\ref{tab:app:mtop_results} presents the same set of results for the six languages in the \mtop{} dataset.

\subsection{Role of decoding strategy and filtering}
\label{sec:app:decoding}
In Table~\ref{tab:app:decoding_and_filtering_appendix} we present results for different decoding strategies and role of filtering inconsistent examples as discussed in Section~\ref{sec:design_choices}.

\begin{table}[h!]
\centering
\begin{adjustbox}{width=\columnwidth}
\begin{tabular}{c cccccc}
\toprule
\textbf{Decoding} & \multirow{2}{*}{\textbf{de}} & \multirow{2}{*}{\textbf{es}} & \multirow{2}{*}{\textbf{fr}} & \multirow{2}{*}{\textbf{hi}} & \multirow{2}{*}{\textbf{th}} & \multirow{2}{*}{\textbf{\average}} \\ 
\textbf{Strategy} &  &  &  &  & &  \\ \midrule
\addlinespace[1mm]
\multicolumn{1}{l}{\textbf{Greedy}}    &  71.1 &	71.7 &	72.6 &	68.1 &	66.0 &  69.9  \\
\multicolumn{1}{l}{\textbf{\;+ Filtering}}  &  72.2 &	73.5	& 74.8 &	71.5 &	67.4 &	71.9  \\
\midrule
\multicolumn{7}{l}{\textbf{{\Topk} Sampling} ($k=40$) }\\  
\small (\#samples)  &  &  &  &  &  &    \\
1 &   70.5	&	71.7  &	73.1  &	66.8  &	66.5   &  69.6  \\
2 &   72.3	&	72.7  &	75.7  &	68.7  &	67.3   &  71.3  \\
4 &   71.3	&	73.1  &	73.8  &	68.5  &	67.8   &  70.9  \\
8 &   71.1	&	72.5  &	74.2  &	69.3  &	67.5   &  70.9  \\ 
\midrule
\multicolumn{7}{l}{\textbf{ \Topk{} Sampling}  + \textbf{Filtering} ($k=40$) } \\
\small (\#samples)  &  &  &  &  &   &  \\
1 &  72.4	 &	74.4 &	78 &	70.9 &	66.1    &	72.4    \\ 
2 &  73.6 &	74.4 &	78.2 &	72.1 &	67.9   &  73.2 \\
4 &  73.4 &	75.3 &	78.8 &	71.4 &	67.1    &	73.2    \\
8 &  73.7 &		75.2 &	\textbf{79.7} &	\textbf{72.5} &	66.8    &	73.6   \\
\midrule
\multicolumn{7}{l}{\textbf{{\Topp} Sampling} ($p=0.95$)} \\  
\small (\#samples)  &  &  &  &  &  &    \\
1 &	70.1  &	71.5 &	74.3 &	66.9 &	67.2 &	70.0 \\
2 &	71.4  &	72.1 &	74.5 &	68.8 &	67.2 &	70.8 \\
4 &	71.1  &	72.8 &	76.4 &	69.0 &	66.0 &	71.1 \\
8 &	71.9  &	72.7 &	74.2 &	70.0 &	68.4 &	71.4 \\
\midrule
\multicolumn{7}{l}{\textbf{\Topp{} Sampling} + \textbf{Filtering} ($p=0.95$)} \\
\small (\#samples)  &  &  &  &  &   &  \\
1  &	72.0 &	75.2 &	78.9 &	71.6 &	\textbf{68.1} &	73.2 \\ 
2  &	73.7 &	75.2 &	79.5 &	72.0 &	67.6 &	73.6 \\ 
4  &	73.4 &	75.3 &	79.0 &	72.1 &	67.7 &	73.5 \\ 
8  &	\textbf{74.0} &	\textbf{75.4} &	79.6 &	72.3 &	68.0 &	\textbf{73.9} \\ \bottomrule

\end{tabular}
\end{adjustbox}
\caption{\textbf{Impact of decoding strategy and filtering}: Generating multiple translations per English example using {\topk} or {\topp} sampling followed by filtering inconsistent examples offers superior downstream performance compared to using greedy decoding or sampling just one translation per example.}
\label{tab:app:decoding_and_filtering_appendix}
\end{table}

\end{document}